\documentclass[letterpaper, 10 pt, conference]{URL-ieeeconf}
\IEEEoverridecommandlockouts    
\usepackage{graphics}           
\usepackage{times}              
\usepackage{amsmath}            
\usepackage{amssymb}            
\usepackage{graphicx}
\usepackage{algorithm}
\usepackage[noend]{algpseudocode}
\usepackage{booktabs}
\usepackage{color}
\usepackage{multirow}
\usepackage{subcaption}
\usepackage{rotating}
\usepackage{cite}
\definecolor{instructioncolor}{rgb}{.5,.5,.5}
\usepackage{soul}

\def\secref#1{Section~\ref{#1}}
\def\figref#1{Fig.~\ref{#1}}
\def\tabref#1{Table~\ref{#1}}
\def\eqref#1{(\ref{#1})}

\def\vstab{\vspace{-0.3cm}}

\newcommand{\rom}[1]{\uppercase\expandafter{\romannumeral #1\relax}}

\makeatletter
\usepackage{xspace}
\DeclareRobustCommand\onedot{\futurelet\@let@token\@onedot}
\def\@onedot{\ifx\@let@token.\else.\null\fi\xspace}

\def\etal{{\textit{et al}}\onedot}
\makeatother

\def\etalcite#1{\etal~\cite{#1}}

\usepackage{array}
\newcolumntype{L}[1]{>{\raggedright\let\newline\\\arraybackslash\hspace{0pt}}m{#1}}
\newcolumntype{C}[1]{>{\centering\let\newline\\\arraybackslash\hspace{0pt}}m{#1}}
\newcolumntype{R}[1]{>{\raggedleft\let\newline\\\arraybackslash\hspace{0pt}}m{#1}}

\def\argmin{\mathop{\rm argmin}}

\title{\LARGE \bf Galibr: Targetless LiDAR-Camera Extrinsic Calibration Method 

via Ground Plane Initialization}

\author{Wonho Song$^{1}$, Minho Oh$^{1}$, Jaeyoung Lee$^{2}$, and Hyun Myung$^{1*}$, \textit{Senior Member, IEEE}
  \thanks{$^*$Corresponding author: Hyun Myung}
  \thanks{$^{1}$Wonho Song, Minho Oh, and Hyun Myung are with the School of Electrical Engineering, KAIST (Korea Advanced Institute of Science and Technology), Daejeon, 34141, Republic of Korea. {\tt\scriptsize \{swh4613, minho.oh, hmyung\}@kaist.ac.kr}} \hfill \break
  \thanks{$^{2}$Jaeyoung Lee is with the Hanwha Aerospace, 6, Pangyo-ro 319 beon-gil, Bundang-gu, Seongnam-si, Gyeonggi-do, 13488, Republic of Korea. {\tt\scriptsize jaeyoung1.lee@hanwha.com} \hfill \break 
  \indent This work has been supported by the grant from Hanwha Aerospace as part of the development of autonomous driving technology for unstructured environments. The students are supported by BK21 FOUR.}
}

\begin{document}
\maketitle
\thispagestyle{empty}
\pagestyle{empty}

\begin{abstract}
  %
  

  With the rapid development of autonomous driving and SLAM technology, the performance of autonomous systems using multimodal sensors highly relies on accurate extrinsic calibration. Addressing the need for a convenient, maintenance-friendly calibration process in any natural environment, this paper introduces Galibr, a fully automatic targetless LiDAR-camera extrinsic calibration tool designed for ground vehicle platforms in any natural setting. The method utilizes the ground planes and edge information from both LiDAR and camera inputs, streamlining the calibration process. It encompasses two main steps: an initial pose estimation algorithm based on ground planes (GP-init), and a refinement phase through edge extraction and matching. Our approach significantly enhances calibration performance, primarily attributed to our novel initial pose estimation method, as demonstrated in unstructured natural environments, including on the KITTI dataset and the KAIST quadruped dataset.

\end{abstract}

\section{Introduction}
\label{sec:intro}




The extrinsic calibration of sensors is a critical component in the advancement of autonomous driving and robotics. Proper sensor fusion, which is essential for these fields, depends heavily on the precise alignment of different sensor modalities. Traditional calibration methods, which rely on target boards, are cumbersome, time-consuming, and ill-suited for dynamic and changing environments. Furthermore, other targetless calibration methods, which may not require physical targets, still often depend on specific environmental features or manual interventions.

\begin{figure}[t]
	\centering
    \includegraphics[page=1, width=1.0\linewidth]{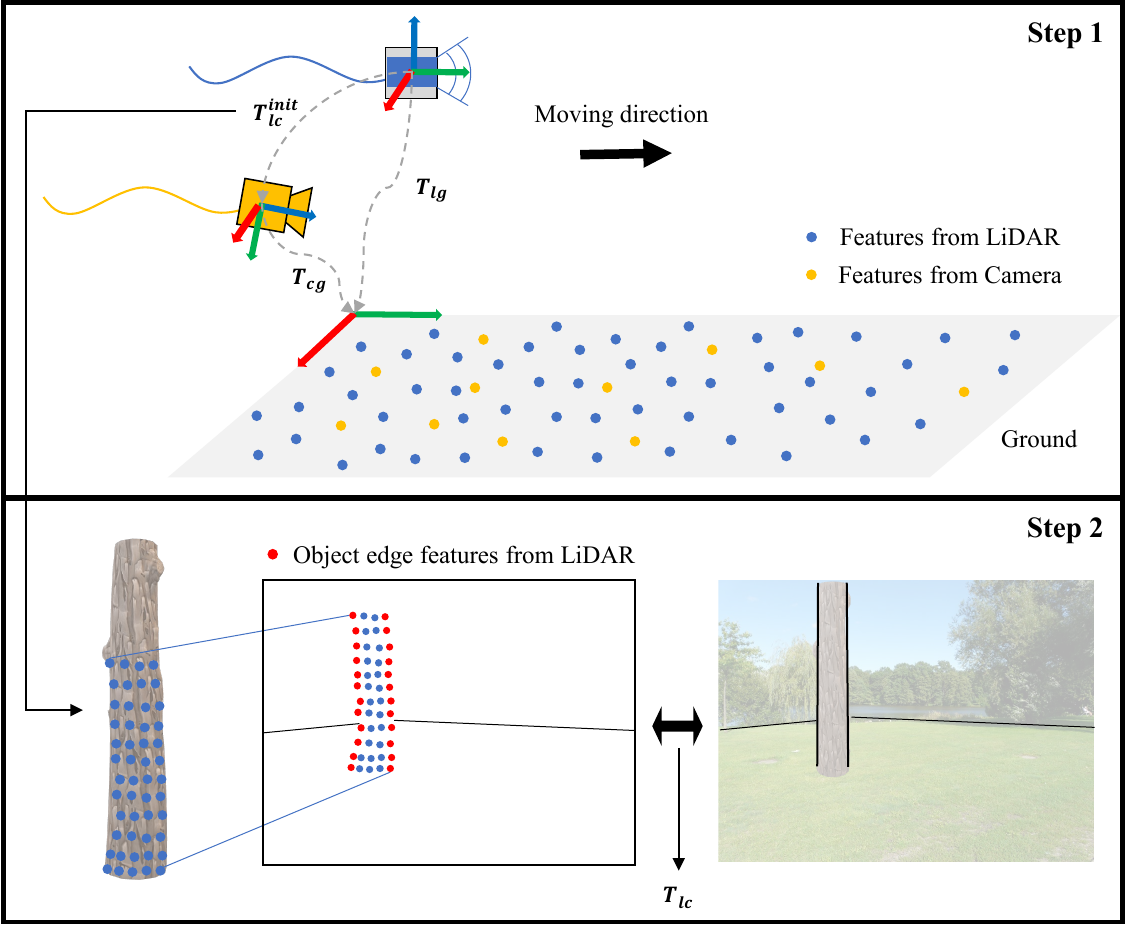}
	\captionsetup{font=footnotesize}
	\caption{Overview of Galibr. Galibr estimates the LiDAR-camera extrinsic calibration result in two steps: initial relative pose estimation using a ground plane and edge matching-based extrinsic calibration.}
	\label{fig:overview}
    \vstab
\end{figure}

This paper introduces a novel, automatic, and targetless calibration method that leverages the stability of the ground plane to provide robust initial values for sensor alignment and refines using edge matching, facilitating a more streamlined and maintenance-friendly approach to calibration. We present the overview of our approach in \figref{fig:overview}. This method is uniquely applicable to a broad range of operational contexts, including challenging terrains. It is designed to be executed in motion, offering a significant improvement in terms of convenience and efficiency. 

The contributions of our approach can be outlined as follows:

\begin{itemize}
    \item We present a novel, fully automatic, and targetless LiDAR-camera extrinsic calibration method, called Galibr, that can be performed in any natural environment. Our approach can be applied to any ground vehicle platform and various sensor setups.
    \item For robust and accurate calibration, we present an initial guess estimation algorithm using ground planes, called GP-init. Relative initial pose estimation from the ground improves the performance of the Galibr system.
    \item We enhance calibration precision through non-ground object edge extraction and matching, capitalizing on the detailed edge data from both LiDAR and camera inputs.
    \item Our method is tested in unstructured natural environments, including the KITTI dataset and the KAIST quadruped dataset, showcasing its superior performance and practical applicability.
\end{itemize}





\section{Related Work}
\label{sec:related}

\subsection{Target-based Calibration Methods}\label{subsec:targetbased}

In the domain of LiDAR-camera extrinsic calibration, target-based approaches are prevalent due to their established precision. Commonly used calibration targets include checkerboard patterns~\cite{geiger2012icra, zhou2018iros} as well as planar boards with various shapes~\cite{velas2014wscg}, and 3D structured objects~\cite{kummerle2020icra, gong2013sensors}. These methods, while precise, often necessitate laborious setup and repeated data collection~\cite{beltran2022tits, fang2021iros}. Recent advances strive to streamline this process by extracting geometric features from point clouds and images, such as corner points~\cite{geiger2012icra} and line features~\cite{zhou2018iros}, without manual intervention. However, the practicality of deploying such target-based methods in real-world applications, particularly in autonomous driving, remains constrained due to the challenges associated with the manual placement of targets and the necessity for stable calibration environments.

\subsection{Target-less Calibration Methods}\label{subsec:targetless}

In targetless LiDAR-camera calibration methods, artificial environmental features like edges and data intensities have been prominent. Techniques such as those by Kang~\etalcite{kang2020jfr}, Yuan~\etalcite{yuan2021ral}, Liu~\etalcite{liu2022tim}, and Chen~\etalcite{chen2023icra} optimize extrinsic parameters by aligning natural edges from LiDAR and camera data. Additionally, mutual information metrics have been employed to align point cloud and image intensities for calibration refinement~\cite{pandey2015jfr}. On the other hand, motion-based methods, exemplified by Taylor~\etalcite{taylor2016tro}, leverage sensor's ego-motion for initial calibration, often requiring significant special sensor movement. Moreover, deep learning based approaches, including the works of Schneider~\etalcite{schneider2017iv}, Iyer~\etalcite{iyer2018iros}, and Koide~\etalcite{koide2023icra}, employ neural networks to extract features and regress extrinsic parameters simultaneously. However, the generalizability of these methods is typically confined to the dataset used for training, limiting their practical application across varying environments.

Building upon the limitations of prior work, our approach, Galibr, is adaptable to any natural environment, including rough terrains, and applicable to all ground-moving platforms. Unlike the reliance on manually defined or environmental-specific features \cite{kang2020jfr,yuan2021ral,liu2022tim,chen2023icra}, Galibr introduces a unique methodology that centers on a novel initial guess estimation utilizing robust ground plane identification, ensuring a more versatile and environment-agnostic initial calibration. Furthermore, Galibr advances the calibration process by implementing a novel edge matching method for the refinement, distinct from the direct feature comparison or mutual information metrics used in prior works \cite{pandey2015jfr,taylor2016tro,schneider2017iv,iyer2018iros,koide2023icra}. This dual-phase process, combining ground plane identification with sophisticated edge matching, marks a significant departure from existing target-less calibration methods by offering enhanced adaptability, accuracy, and applicability across diverse operational environments. The overview of the Galibr and its detailed steps are presented in \secref{sec:main}.


\begin{figure*}[t]
	\centering
    \includegraphics[page=3, width=1.0\textwidth]{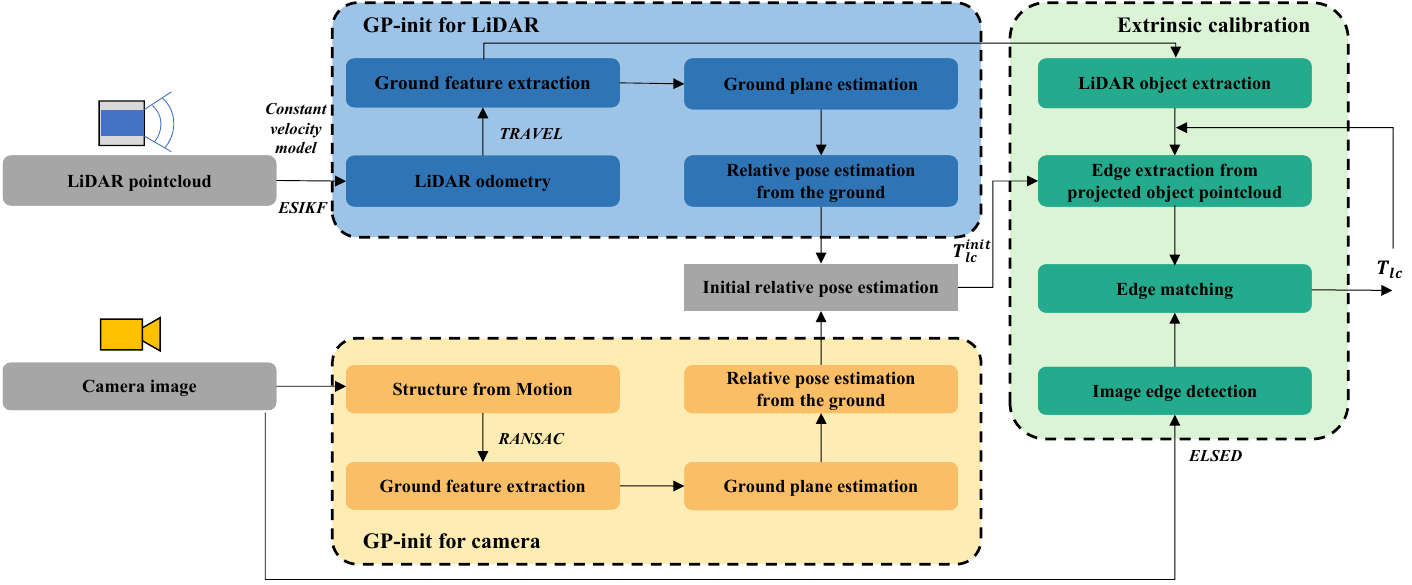}
	\captionsetup{font=footnotesize}
	\caption{System overview of our approach. Unlike existing LiDAR and camera extrinsic calibration methods, which generally need initial values, our approach focuses on estimating initial relative pose using ground plane features. With the two-step estimation, initial pose estimation step and extrinsic calibration step, our approach outputs more accurate and robust extrinsic calibration results.}
	\label{fig:framework}
    \vstab
\end{figure*}

\section{Galibr: Target-less Extrinsic Calibration Method via Ground Plane Initialization}
\label{sec:main}

Galibr's calibration procedure requires the platform to be in motion in any environment to estimate each sensor's pose. The overall system of Galibr is structured by two main steps: the initial pose estimation step utilizing ground plane, called GP-init, and the extrinsic calibration step using edge matching as presented in \figref{fig:framework}. GP-init contains the ground feature extraction module and the relative pose estimation module from the ground for each sensor. Extrinsic calibration is performed to match edges extracted from each sensor. 



\subsection{GP-init for Camera}\label{subsec:camgpinit}

We describe our approach for ground vehicle pose estimation and ground plane extraction, integrating Structure from Motion (SfM), RANSAC~\cite{fischler1981comm}-based ground plane estimation, vertical alignment, and relative pose estimation.

During this process, we employed SfM to extract 3D ground features, which are crucial for estimating the camera's ego-motion. The motion direction vector \( v^c \) is derived from the normalized incremental position vector \( \Delta p^c \), scaled by \( s \):
\begin{equation}
    v^c = \frac{s \cdot \Delta p^c}{\| \Delta p^c \|} .
\end{equation}

For ground plane extraction in camera coordinates, we utilized the 3D ground features estimated via SfM, exploiting the RANSAC algorithm. The optimal plane parameters \( n^c = [a^c, b^c, c^c, d^c] \) are found by maximizing the inlier consensus set.


To align robot motion with the ground plane, we find the normal vector of ground plane \( n^c \) over a sliding window from time \( t_1^c \) to \( t_2^c \) using a threshold \( \epsilon^c \):
\begin{equation}
    \sum_{t=t_1^c}^{t_2^c} \left(v^c \cdot n^c\right)_{t} \leq \epsilon^c .
\end{equation}

This condition ensures that the motion of the camera system is closely aligned with the ground plane. It is essential for identifying a planar ground by ensuring the dot product between the camera's direction vector \(v^c\) and the ground plane's normal vector \(n^c\) remains below a certain threshold. This approach helps confirm the ground plane's planarity, indicating that the camera’s motion aligns well with a stable, flat surface.

The relative pose between the ground and camera, \( T^c_g \), includes scaled translation \( z^c \) and orientation angles \( \theta_{\text{roll}}^c \) and \( \theta_{\text{pitch}}^c \):
\begin{align}
    z^c &= s \cdot d^c, \\
    \theta_{\text{roll}}^c &= \arctan2\left(-b^c, c^c\right), \\
    \theta_{\text{pitch}}^c &= \arctan2\left(-a^c, \sqrt{{b^c}^2 + {c^c}^2}\right) .
\end{align}

The pose \( T^c_g \) is represented as \( (R^c_g, p^c_g) \), where \( R^c_g \) is the rotation matrix and \( p^c_g \) is the position vector from the ground to the camera.

\subsection{GP-init for LiDAR}\label{subsec:lidargpinit}

\begin{figure}[t]
    \centering
    \captionsetup{font=footnotesize}
    \vspace{2mm}
    \begin{subfigure}[b]{0.23\textwidth}
        \includegraphics[page=8, width=1.0\linewidth]{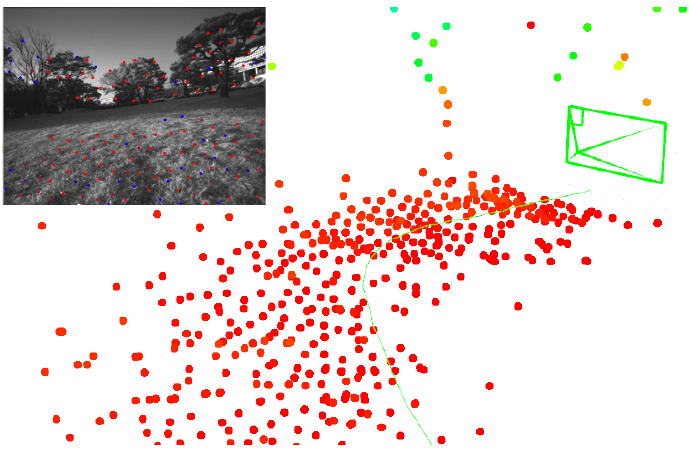}
        \caption{}
    \end{subfigure}
    \begin{subfigure}[b]{0.23\textwidth}
        \includegraphics[page=9, width=1.0\linewidth]{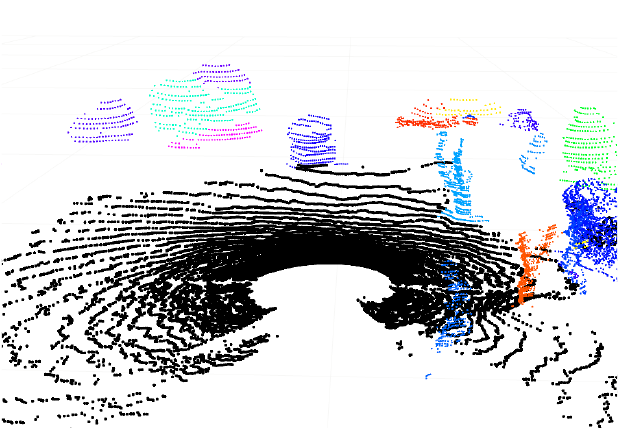}
        \caption{}
    \end{subfigure}
    \caption{(a) Ground feature extraction result (red points) using SfM. (b) Ground feature extraction result (black points) using LiDAR odometry and TRAVEL~\cite{oh2022ral}. Using ground extraction results, we estimate the ground plane and the relative pose from the ground for the initial relative pose estimation of two sensors.}
	\label{fig:ground}
    \vstab
\end{figure}

In the methodology for estimating the relative pose of the LiDAR with respect to the ground, the LiDAR odometry estimation is initiated by the Iterated Error State Kalman Filter (IESKF) with a constant velocity model. Ground feature extraction follows, utilizing the TRAVEL~\cite{oh2022ral} method which is the state-of-the-art ground and non-ground segmentation method, to distinguish ground from non-ground point clouds and cluster non-ground objects. The ground plane is estimated using the RANSAC~\cite{fischler1981comm} algorithm, ensuring alignment of the LiDAR's motion direction with the ground plane. This step involves maintaining the dot product of the ground plane's normal vector and the direction vector from LiDAR odometry below a predefined threshold over a sliding window, as described in \secref{subsec:camgpinit}:
\begin{equation}
    \sum_{t=t_1^l}^{t_2^l} \left(v^l \cdot n^l\right)_{t} \leq \epsilon^l, \quad  v^l = \frac{\Delta p^l}{\| \Delta p^l \|} ,
\end{equation}
where \( v^l \) is the motion direction vector of a LiDAR, \( n^l = [a^l, b^l, c^l, d^l] \) are the parameters of the ground plane in LiDAR coordinates, \( \Delta p^l \) is the incremental position vector of the LiDAR, and \( \epsilon^l \) represents the alignment threshold. To ensure that the motion of the LiDAR system is closely aligned with the ground plane, we find the normal vector of ground plane \( n^l \) over a sliding window from time \( t_1^l \) to \( t_2^l \) using a threshold \( \epsilon^l \). The ground feature extraction result from a LiDAR using TRAVEL~\cite{oh2022ral} is presented in \figref{fig:ground} (b).

The process concludes with relative pose estimation, akin to the approach in \secref{subsec:camgpinit}, where the LiDAR's pose relative to the ground plane is determined. The transformation matrix \( T^l_g \) representing the LiDAR's pose in relation to the ground is composed of the rotation from the ground to the LiDAR \( R^l_g \) and the position from the ground to the LiDAR \( p^l_g \):
\begin{equation}
    T^l_g = (R^l_g, p^l_g) .
\end{equation}
Here, \( R^l_g \) and \( p^l_g \) reflect the rotational and positional components from the ground to the LiDAR, respectively.

As a result, the GP-init phase provides an initial guess that includes estimating each sensor's height above the ground and the orientation angles, specifically the roll and pitch. Our methodology combines these components for accurate pose estimation of both sensors and ground plane extraction.

\subsection{Extrinsic Calibration Using Edge Matching}\label{subsec:edgematch}

Given the transformation matrices from the ground to the camera \( T^c_g \) and from the ground to the LiDAR \( T^l_g \), the initial relative pose from the LiDAR to the camera, denoted as \( T^l_c \), can be estimated. This is achieved by first inverting the transformation from the ground to the LiDAR to obtain the transformation from the LiDAR to the ground, and then concatenating it with the transformation from the ground to the camera. Mathematically, this can be expressed as:

\begin{equation}
    T_l^c = (T^l_g)^{-1} T^c_g .
\end{equation}

This operation effectively chains the relative transformations, converting the pose from the ground frame to the LiDAR frame and then from the LiDAR frame to the camera frame, yielding the desired initial relative pose from the LiDAR to the camera.

The initial relative pose estimation involves the computation of the transformation matrix \( T^l_c \), representing the pose of the LiDAR relative to the camera. Given the transformation matrices \( T^l_g \) and \( T^c_g \) from the LiDAR and camera to the ground respectively, \( T_l^c \) can be derived as follows:
\begin{equation}
    T^l_c = (T^c_g)^{-1} T^l_g ,
\end{equation}
where \( (T^c_g)^{-1} \) denotes the inverse of the transformation matrix from the camera to the ground, effectively repositioning the reference from the ground to the camera frame.

\begin{figure}[t]
    \centering
    \vspace{2mm}
    \includegraphics[page=4, width=1.0\linewidth]{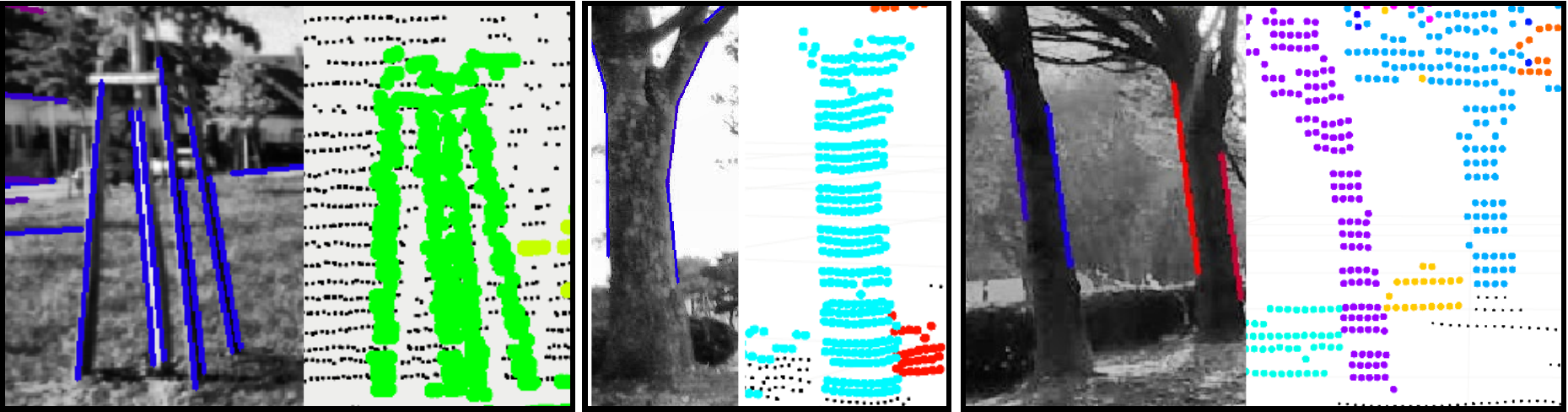}
    \captionsetup{font=footnotesize}
    \caption{The edge detection examples in images and object extraction examples in LiDAR point clouds.}
    \label{fig:edge}
    \vstab
\end{figure}

\begin{figure}[t]
    \centering
    \vspace{2mm}
    \includegraphics[page=2, width=0.9\linewidth]{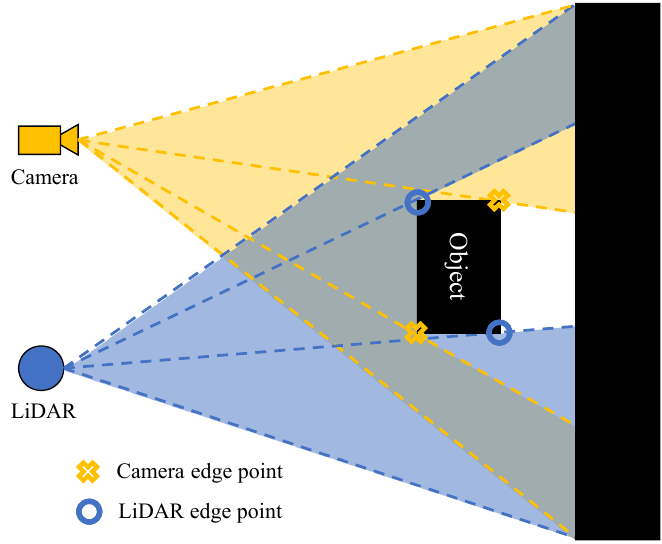}
    \captionsetup{font=footnotesize}
    \caption{Visual description of different edge points of a foreground object from different views of LiDAR and camera.}
    \label{fig:object}
    \vstab
\end{figure}

For edge extraction from the LiDAR point cloud and the image, ELSED~\cite{suarez2022elsed} which is the fastest existing edge extraction method, is employed for image edge detection, and non-ground segmented point clouds are used for LiDAR edge detection. The edge detection examples are presented in \figref{fig:edge}. The LiDAR non-ground objects are projected onto the image plane using the initial relative pose \( T^l_c \), facilitating the identification of edge points. When extracting edges from the LiDAR pointcloud, it is essential to note that the actual edges observed in the image may not align perfectly with the LiDAR-detected edges, as illustrated in \figref{fig:object}. This discrepancy arises because most of the existing methods using edges for extrinsic calibration\cite{zhang2021icra}\cite{castorena2016icassp}\cite{wang2023iros} utilize LiDAR edges from LiDAR view, which may not correspond directly to the real image edges. To mitigate this, the initial relative pose \( T^l_c \) is employed to project LiDAR non-ground objects onto the image plane, enabling a more accurate alignment and comparison between LiDAR-detected edges and actual image edges. The difference in edge representation emphasizes the importance of carefully calibrating the relative pose between the LiDAR and camera to ensure the fidelity of edge-based features across both modalities. The projection is formalized as:
\begin{equation}
    X^c = K (T^l_c X^l) ,
\end{equation}
where \( X^c \) and \( X^l \) denote the homogeneous coordinates of points in the image and LiDAR frames, respectively, and \( K \) represents the camera's intrinsic matrix.

In the extrinsic calibration optimization process, the goal is to fine-tune the initial transformation matrix \( T^l_c \) that aligns the LiDAR and camera frames. We minimize the reprojection error of the 3D occlusion edge points \( E_i^l \) onto the corresponding 2D edge lines \( E_i^c \) in the image. The optimization problem is formulated as a Perspective-n-Point (PnP) problem and is expressed as:

\begin{equation}
    \xi^* = \argmin_{\xi} \sum_{i} \rho_i \left( \left\| f_r \left( \pi \left( \exp(\xi) \circ T^l_c \circ E_i^l \right), E_i^c \right) \right\| \right)^2 .
\end{equation}

Here, \( \xi \) represents the Lie algebra elements corresponding to the extrinsic parameters we aim to optimize, \( f_r \) denotes the perpendicular distance function from a point to a line, and \( \rho_i \) is a robust cost function such as the Huber kernel, which helps mitigate the influence of outliers. The projection \( \pi \) maps 3D LiDAR points onto the 2D image plane using the initial guess \(T^l_c \) and the updated transformation \( \exp(\xi) \). The Levenberg-Marquardt (LM) algorithm is utilized to solve this optimization iteratively, refining the estimate for \( \xi \), until convergence criteria are met. The outcome is the optimized extrinsic calibration matrix \( {T^l_c}^* = \exp(\xi^*) \circ T^l_c \), which represents the most accurate transformation between the LiDAR and camera coordinate systems.

\section{Experimental Evaluation}
\label{sec:exp}

This study primarily emphasizes the development of an automatic and robust extrinsic calibration method in unstructured natural environments. Accordingly, we present our experiments to support our key claims that our approach (i) shows promising performance compared with state-of-the-art methods in unstructured natural environments, (ii) improves performance when using GP-init, and (iii) operates at a faster speed than other extrinsic calibration methods. Our experimental evaluation backs up these claims in this section.

\subsection{Experimental Setup}
To evaluate the extrinsic calibration performance, we used the KITTI dataset \cite{geiger2012cvpr}, which includes an accurate ground truth calibration matrix. Furthermore, we also tested our methods on our own dataset, called KAIST quadruped dataset, which was collected by using a quadruped robot, including Ouster OS0-128 and Intel RealSense D435i. The ground truth of the extrinsic calibration between the LiDAR and the camera was obtained using the same methodology as described in \cite{lee2022ral}. We collected the KAIST quadruped dataset in unstructured outdoor environments with rough terrains on the KAIST campus.




\begin{figure}[t]
    \centering
    \vspace{2mm}
    \includegraphics[page=7, width=1.0\linewidth]{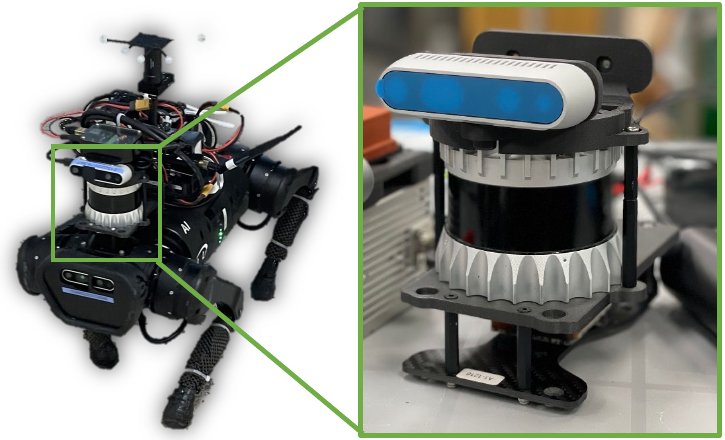}
    \captionsetup{font=footnotesize}
    \caption{Sensor setup of a quadruped robot platform for the KAIST quadruped dataset.}
    \label{fig:experiment}
    \vstab
\end{figure}

\subsection{Performance Comparison with the State-of-the-Art Methods}

We first evaluate the performance of our approach and baseline methods, namely, Liu \cite{liu2022tim}, Chen \cite{chen2023icra}, and Koide \cite{koide2023icra}, to support the claim that our approach shows promising performance compared with other state-of-the-art methods in unstructured natural environments. Particularly, for Koide \cite{koide2023icra}, a deep learning method, we utilized the same weights that were trained on the MegaDepth dataset, ensuring a fair comparison across methodologies. These approaches generally present precise extrinsic performance, but these methods struggle with unstructured environments with rough terrain without close artificial structures. Additionally, these approaches are well suited to solid-state LiDARs, which provide a very dense point cloud in a certain view. We evaluated these methods with traditional spinning LiDARs with the accumulated point cloud using LiDAR-only odometry, which was used in our proposed approach. The other state-of-the-art approaches showed large errors as presented in \tabref{table:accuracy}. Only Koide \cite{koide2023icra} succeeded in calculating the result in the KAIST quadruped dataset. The other state-of-the-art methods failed to estimate the extrinsic calibration matrix in the KAIST quadruped dataset. In contrast, our proposed method using GP-init succeeded and achieved the highest accuracy in both datasets.

\begin{table}[t!]
    \centering
    \vspace{2mm}
    \captionsetup{font=footnotesize}
    \caption{Comparison of calibration errors with the state-of-the-art methods on the KITTI dataset \cite{geiger2012cvpr} and the KAIST quadruped dataset.}
    \renewcommand{\arraystretch}{1.5} 
    \setlength\tabcolsep{3.5 pt} 
    {\scriptsize
        \begin{tabular}{@{}cccccccc@{}}
            \toprule \midrule
            \multirow{2}{*}{Dataset} & \multirow{2}{*}{Method} & \multicolumn{3}{c}{Translation error [cm]} & \multicolumn{3}{c}{Rotation error [°]} \\ \cline{3-8} 
             &  & \hspace*{1.5mm} \textit{x} \hspace*{1.5mm} & \hspace*{1.5mm} \textit{y} \hspace*{1.5mm} & \hspace*{1.5mm} \textit{z} \hspace*{1.5mm} & Roll & Pitch & Yaw  \\ \midrule
            \multirow{5}{*}{\begin{tabular}[c]{@{}c@{}}KITTI dataset \\ (Seq. 06) \cite{geiger2012cvpr} \end{tabular}} & Liu \cite{liu2022tim} & 8.70 & 8.79 & 10.30 & 1.86 & 2.59 & 1.12 \\ \cline{2-8} 
             & Chen \cite{chen2023icra} & 5.69 & 6.36 & 5.84 & 1.65 & 1.10 & 0.85 \\ \cline{2-8} 
             & Koide \cite{koide2023icra} & 5.74 & 4.53 & 5.98 & 0.97 & 0.89 & 0.91 \\ \cline{2-8} 
             & Ours (w/o GP-init) & 5.82 & 4.98 & 6.83 & 1.09 & 1.13 & 0.97 \\ \cline{2-8} 
             & \textbf{Ours (w/ GP-init)} & \textbf{4.32} & \textbf{4.03} & \textbf{1.80} & \textbf{0.83} & \textbf{0.65} & \textbf{0.70} \\ \midrule
            \multirow{3}{*}{\begin{tabular}[c]{@{}c@{}} KAIST quadruped \\ dataset \end{tabular}} & Koide \cite{koide2023icra} & 4.34 & 3.38 & 3.95 & 1.03 & 1.22 & 1.27 \\ \cline{2-8} 
            & Ours (w/o GP-init) & 5.39 & 3.17 & 5.90 & 0.92 & 1.34 & 1.12 \\ \cline{2-8} 
            & \textbf{Ours (w/ GP-init)} & \textbf{1.90} & \textbf{1.02} & \textbf{0.72} & \textbf{0.56} & \textbf{0.54} & \textbf{0.93} \\ \midrule
            \bottomrule
        \end{tabular}
	}
	\label{table:accuracy}
    \vstab
\end{table}


\subsection{Performance Improvement Using GP-init}
\label{subsec:evalgpinit}

The second experiment evaluates in two settings: GP-init and without GP-init. It illustrates that our approach is capable of improving performance when using the initial pose estimation step. GP-init helped to estimate accurate and robust initial values, and it improved the accuracy of the result as presented in \tabref{table:accuracy}. The \textit{z-}axis, roll, and pitch errors have significantly reduced when using GP-init. 

The visualization results are also presented in Figs. 7 and 8. The figures show fused images integrating non-ground LiDAR point cloud data onto the corresponding camera image. The first rows present the extrinsic calibration result of our approach without using GP-init, and the second rows show the result with GP-init. The non-ground objects are somewhat misaligned when GP-init is not performed. When two steps of our approach are all performed, the visualization result shows highly aligned results in both datasets.

\begin{figure*}[t]
    \centering
    \vspace{2mm}
    \includegraphics[page=5, width=1.0\linewidth]{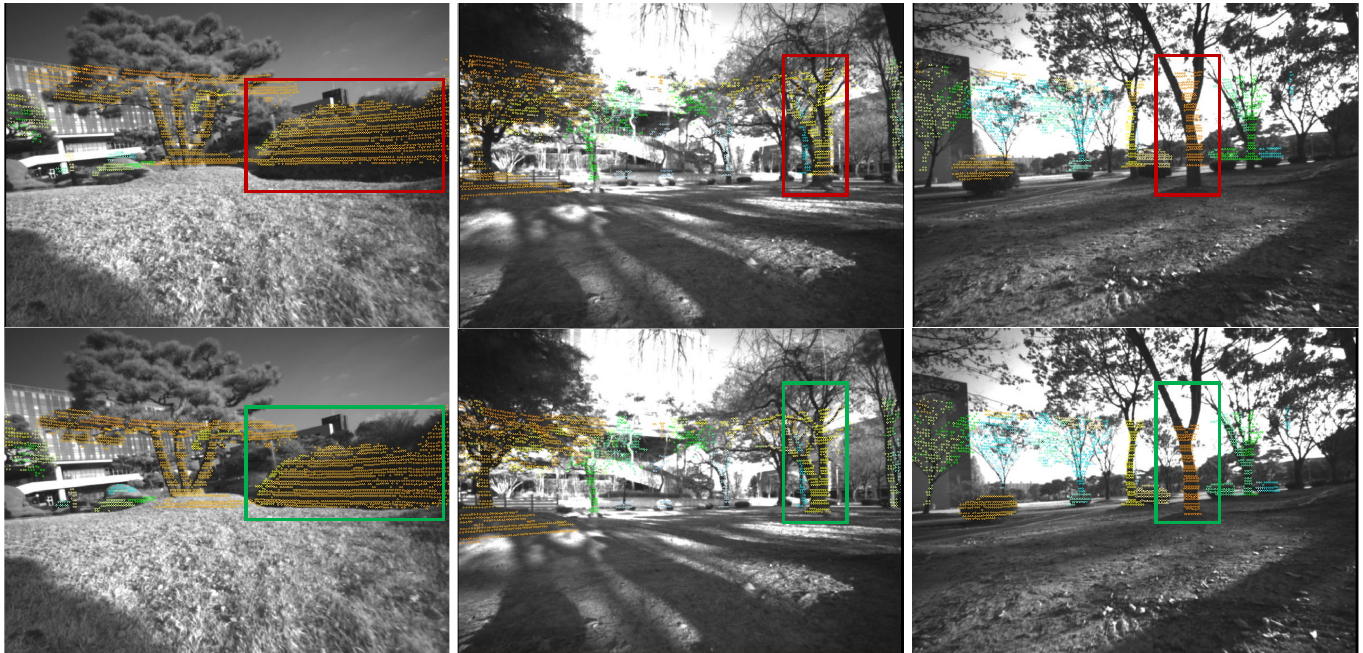}
    \captionsetup{font=footnotesize}
    \caption{The experimental result comparing two methods with the KAIST quadruped dataset. (Top) LiDAR non-ground point cloud projected onto the camera image using a mis-calibrated result when GP-init was not used. (Bottom) The LiDAR point cloud projected image using result of our approach with GP-init.}
	\label{fig:result_keit}
    \vstab
\end{figure*}

\begin{figure*}[t]
    \centering
    \includegraphics[page=6, width=1.0\linewidth]{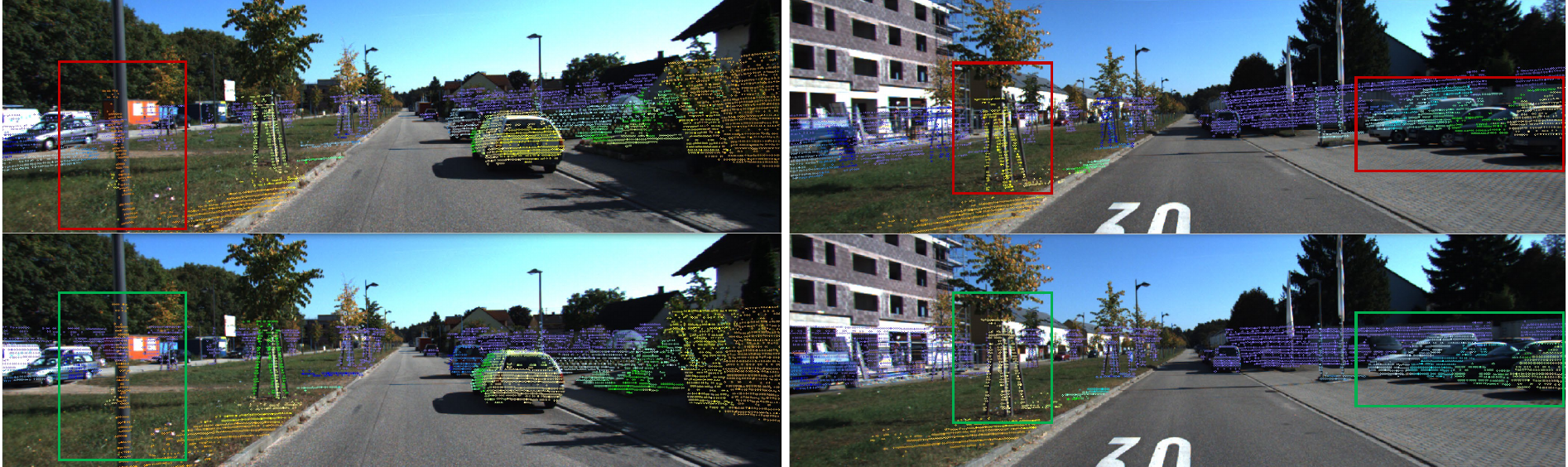}
    \captionsetup{font=footnotesize}
    \caption{The experimental result comparing two methods with the KITTI dataset \cite{geiger2012cvpr}. (Top) LiDAR non-ground point cloud projection results onto the camera image using a miscalibrated result when GP-init was not used. (Bottom) LiDAR non-ground point cloud projection results from our approach with GP-init.}
	\label{fig:result_kitti}
    \vstab
\end{figure*}

\subsection{Runtime}

The next set of experiments has been conducted to support the claim that our approach runs faster than other existing LiDAR and camera extrinsic calibration methods. We, therefore, tested our approach on the KITTI dataset \cite{geiger2012cvpr} and checked the processing time of the initial guess step, calibration step, and total. \tabref{table:runtime} summarizes the runtime results for our approach and other state-of-the-art methods. The numbers support our third claim, namely that the computations can be executed fast. With the same condition on the single-thread mode using AMD Ryzen 9 5950X CPU, we achieved the fastest speed with the fastest average initial guess and calibration steps.

\begin{table}[t]
    \centering
    \vspace{2mm}
    \captionsetup{font=footnotesize}
    \caption{Mean computation speed of extrinsic calibration in the KITTI dataset. The speed of each algorithm was fairly checked on the single-thread mode using AMD Ryzen 9 5950X CPU.}
    \renewcommand{\arraystretch}{1.4} 
    \setlength\tabcolsep{12.1 pt} 
    {\scriptsize
        \begin{tabular}{@{}ccccc@{}}
            \toprule \midrule
            \multirow{2}{*}{Dataset} & \multirow{2}{*}{Method} & \multicolumn{3}{c}{Processing time [s]} \\ \cline{3-5}
            & & Init. guess & Calibration & Total \\ \midrule
            \multirow{4}{*}{\begin{tabular}[c]{@{}c@{}}KITTI dataset \\ (Seq. 06) \cite{geiger2012cvpr} \end{tabular}} & Liu \cite{liu2022tim} & 8.28 & 0.91 & 9.19  \\ \cline{2-5}
            & Chen \cite{chen2023icra} & 60.28 & 15.74 & 76.02  \\ \cline{2-5}
            & Koide \cite{koide2023icra} & 11.34 & 32.24 & 43.58  \\ \cline{2-5}
            & \textbf{Ours} & \textbf{5.91} & \textbf{0.85} & \textbf{6.76} \\ \midrule
            \bottomrule
        \end{tabular}
	}
	\label{table:runtime}
    \vstab
\end{table}




\section{Conclusion}
\label{sec:conclusion}

This paper proposes a fast and robust extrinsic calibration method with ground plane-based initialization, showing its robustness in any environment and any ground vehicle platform. However, it still strongly relies on ground feature extraction and edge extraction results. Therefore, in future work, we plan to combine our approach with an IMU sensor to calibrate the multi-modal sensors to establish more generalized and robust automatic extrinsic calibration. Furthermore, we aim to enhance the motion estimation capabilities of the LiDAR and camera by potentially employing advanced techniques such as adaptive filtering, as discussed in Lim~\etalcite{lim2023arxiv}, and refining the matching technique as presented in Lim~\etalcite{lim2023ijrr} and Lim~\etalcite{lim2022ral}, to improve the overall performance of our calibration method.

\bibliographystyle{URL-IEEEtrans}

\bibliography{URL-bib}

\begin{thebibliography}{10}
\providecommand{\url}[1]{#1}
\csname url@rmstyle\endcsname
\providecommand{\newblock}{\relax}
\providecommand{\bibinfo}[2]{#2}
\providecommand\BIBentrySTDinterwordspacing{\spaceskip=0pt\relax}
\providecommand\BIBentryALTinterwordstretchfactor{4}
\providecommand\BIBentryALTinterwordspacing{\spaceskip=\fontdimen2\font plus
\BIBentryALTinterwordstretchfactor\fontdimen3\font minus \fontdimen4\font\relax}
\providecommand\BIBforeignlanguage[2]{{%
\expandafter\ifx\csname l@#1\endcsname\relax
\typeout{** WARNING: IEEEtran.bst: No hyphenation pattern has been}%
\typeout{** loaded for the language `#1'. Using the pattern for}%
\typeout{** the default language instead.}%
\else
\language=\csname l@#1\endcsname
\fi
#2}}

\bibitem{geiger2012icra}
A.~Geiger, F.~Moosmann, O.~Car, and B.~Schuster, ``Automatic camera and range sensor calibration using a single shot,'' in \emph{Proc. IEEE Int. Conf. Robot. Automat.}, 2012, pp. 3936--3943.

\bibitem{zhou2018iros}
L.~Zhou, Z.~Li, and M.~Kaess, ``Automatic extrinsic calibration of a camera and a {3D} {LiDAR} using line and plane correspondences,'' in \emph{Proc. IEEE/RSJ Int. Conf. Intell. Robot. Syst.}, 2018, pp. 5562--5569.

\bibitem{velas2014wscg}
M.~Velas, M.~Spanel, Z.~Materna, and A.~Herout, ``Calibration of {RGB} camera with velodyne {LiDAR},'' in \emph{Proc. Int. Conf. Comput. Graph. Visualizat. Comput. Vis.}, 2014.

\bibitem{kummerle2020icra}
J.~Kümmerle and T.~Kühner, ``Unified intrinsic and extrinsic camera and {LiDAR} calibration under uncertainties,'' in \emph{Proc. IEEE Int. Conf. Robot. Automat.}, 2020, pp. 6028--6034.

\bibitem{gong2013sensors}
X.~Gong, Y.~Lin, and J.~Liu, ``{3D} {LiDAR}-camera extrinsic calibration using an arbitrary trihedron,'' \emph{Sensors}, vol.~13, no.~2, pp. 1902--1918, 2013.

\bibitem{beltran2022tits}
J.~Beltrán, C.~Guindel, A.~d.~l. Escalera, and F.~García, ``Automatic extrinsic calibration method for {LiDAR} and camera sensor setups,'' \emph{IEEE Trans. Intell. Transport. Syst.}, vol.~23, no.~10, pp. 17\,677--17\,689, 2022.

\bibitem{fang2021iros}
C.~Fang, S.~Ding, Z.~Dong, H.~Li, Z.~Siyu, and P.~Tan, ``{Singleshot is enough}: Panoramic infrastructure based calibration of multiple cameras and {3D} {LiDAR}s,'' in \emph{Proc. IEEE/RSJ Int. Conf. Intell. Robot. Syst.}, 2021, pp. 8890--8897.

\bibitem{kang2020jfr}
J.~Kang and N.~L. Doh, ``Automatic targetless camera--{LiDAR} calibration by aligning edge with gaussian mixture model,'' \emph{J. Field Robot.}, vol.~37, no.~1, pp. 158--179, 2020.

\bibitem{yuan2021ral}
C.~Yuan, X.~Liu, X.~Hong, and F.~Zhang, ``Pixel-level extrinsic self-calibration of high-resolution {LiDAR} and camera in targetless environments,'' \emph{IEEE Robot. Automat. Lett.}, vol.~6, no.~4, pp. 7517--7524, 2021.

\bibitem{liu2022tim}
X.~Liu, C.~Yuan, and F.~Zhang, ``Targetless extrinsic calibration of multiple small {FoV} {LiDAR}s and cameras using adaptive voxelization,'' \emph{IEEE Trans. Instrum. Meas.}, vol.~71, pp. 1--12, 2022.

\bibitem{chen2023icra}
F.~Chen, L.~Li, S.~Zhang, J.~Wu, and WangLujia, ``{PBACalib}: Targetless extrinsic calibration for high-resolution {LiDAR}-camera system based on plane-constrained bundle adjustment,'' in \emph{Proc. IEEE Int. Conf. Robot. Automat.}, 2023.

\bibitem{pandey2015jfr}
G.~Pandey, J.~R. McBride, S.~Savarese, and R.~M. Eustice, ``Automatic extrinsic calibration of vision and {LiDAR} by maximizing mutual information,'' \emph{J. Field Robot.}, vol.~32, no.~5, pp. 696--722, 2015.

\bibitem{taylor2016tro}
Z.~Taylor and J.~Nieto, ``Motion-based calibration of multimodal sensor extrinsics and timing offset estimation,'' \emph{IEEE Trans. Robot.}, vol.~32, no.~5, pp. 1215--1229, 2016.

\bibitem{schneider2017iv}
N.~Schneider, F.~Piewak, C.~Stiller, and U.~Franke, ``Regnet: Multimodal sensor registration using deep neural networks,'' in \emph{Proc. IEEE Intell. Veh. Symp.}, 2017, pp. 1803--1810.

\bibitem{iyer2018iros}
G.~Iyer, R.~K. Ram, J.~K. Murthy, and K.~M. Krishna, ``Calibnet: Geometrically supervised extrinsic calibration using {3D} spatial transformer networks,'' in \emph{Proc. IEEE/RSJ Int. Conf. Intell. Robot. Syst.}, 2018, pp. 1110--1117.

\bibitem{koide2023icra}
K.~Koide, S.~Oishi, M.~Yokozuka, and A.~Banno, ``General, single-shot, target-less, and automatic {LiDAR}-camera extrinsic calibration toolbox,'' in \emph{Proc. IEEE Int. Conf. Robot. Automat.}, 2023.

\bibitem{fischler1981comm}
M.~A. Fischler and R.~C. Bolles, ``{Random sample consensus}: A paradigm for model fitting with applications to image analysis and automated cartography,'' \emph{Commun. ACM}, vol.~24, no.~6, pp. 381--395, 1981.

\bibitem{oh2022ral}
M.~Oh, E.~Jung, H.~Lim, W.~Song, S.~Hu, E.~M. Lee, J.~Park, J.~Kim, J.~Lee, and H.~Myung, ``{TRAVEL}: Traversable ground and above-ground object segmentation using graph representation of {3D} {LiDAR} scans,'' \emph{IEEE Robot. Automat. Lett.}, vol.~7, no.~3, pp. 7255--7262, 2022.

\bibitem{suarez2022elsed}
I.~Su{\'a}rez, J.~M. Buenaposada, and L.~Baumela, ``{ELSED}: Enhanced line segment drawing,'' \emph{Pattern Recognit.}, vol. 127, p. 108619, 2022.

\bibitem{zhang2021icra}
X.~Zhang, S.~Zhu, S.~Guo, J.~Li, and H.~Liu, ``Line-based automatic extrinsic calibration of {LiDAR} and camera,'' in \emph{Proc. IEEE Int. Conf. Robot. Automat.}, 2021, pp. 9347--9353.

\bibitem{castorena2016icassp}
J.~Castorena, U.~S. Kamilov, and P.~T. Boufounos, ``Autocalibration of {LiDAR} and optical cameras via edge alignment,'' in \emph{Proc. IEEE Int. Conf. Acoustics, Speech and Signal Process.}, 2016, pp. 2862--2866.

\bibitem{wang2023iros}
S.~Wang, S.~Zhang, and X.~Qiu, ``{P2O-Calib}: Camera-{LiDAR} calibration using point-pair spatial occlusion relationship,'' in \emph{Proc. IEEE/RSJ Int. Conf. Intell. Robot. Syst.}, 2023, pp. 1840--1847.

\bibitem{geiger2012cvpr}
A.~Geiger, P.~Lenz, and R.~Urtasun, ``{Are we ready for autonomous driving? the KITTI vision benchmark suite},'' in \emph{Proc. IEEE/CVF Conf. Comput. Vis. Pattern Recognit.}, 2012, pp. 3354--3361.

\bibitem{lee2022ral}
A.~J. Lee, Y.~Cho, Y.~S. Shin, A.~Kim, and H.~Myung, ``{ViViD++}: Vision for visibility dataset,'' \emph{IEEE Robot. Automat. Lett.}, vol.~7, no.~3, pp. 6282--6289, 2022.

\bibitem{lim2023arxiv}
H.~Lim, D.~Kim, B.~Kim, and H.~Myung, ``{AdaLIO}: Robust adaptive {LiDAR}-inertial odometry in degenerate indoor environments,'' \emph{Proc. Int. Conf. Ubiquti. Robot.}, 2023.

\bibitem{lim2023ijrr}
H.~Lim, B.~Kim, D.~Kim, E.~Mason~Lee, and H.~Myung, ``{Quatro++}: Robust global registration exploiting ground segmentation for loop closing in {LiDAR} {SLAM},'' \emph{Int. J. Robot. Res.}, 2023.

\bibitem{lim2022ral}
H.~Lim, J.~Jeon, and H.~Myung, ``{UV-SLAM}: Unconstrained line-based {SLAM} using vanishing points for structural mapping,'' \emph{IEEE Robot. Automat. Lett.}, vol.~7, no.~2, pp. 1518--1525, 2022.

\end{thebibliography}

\end{document}